\documentclass[ngerman,english]{bvm} 

\bvmdef\articlenumber{3033}

\bvmdef\type{P}

\date{}

\title{2-D Respiration Navigation Framework for 3-D Continuous Cardiac Magnetic Resonance Imaging}

\titlerunning{2-D Respiration Navigation for Cardiac MRI}

\author{Elisabeth~Hoppe$^1$, Jens~Wetzl$^2$, Philipp~Roser$^1$, Lina~Felsner$^1$, Alexander~Preuhs$^1$, Andreas~Maier$^1$}

\authorrunning{Hoppe et al.}

\institute{%
$^1$Pattern Recognition Lab, Friedrich-Alexander-Universit\"at Erlangen-N\"urnberg, Germany\\
$^2$Magnetic Resonance, Siemens Healthineers, Erlangen, Germany}

\email{elisabeth.hoppe@fau.de}

\begin{document}

%
\selectlanguage{english}

\maketitle

\begin{abstract}
	Continuous protocols for cardiac magnetic resonance imaging enable sampling of the cardiac anatomy simultaneously resolved into cardiac phases. To avoid respiration artifacts, associated motion during the scan has to be compensated for during reconstruction. In this paper, we propose a sampling adaption to acquire 2-D respiration information during a continuous scan. Further, we develop a pipeline to extract the different respiration states from the acquired signals, which are used to reconstruct data from one respiration phase. Our results show the benefit of the proposed workflow on the image quality compared to no respiration compensation, as well as a previous 1-D respiration navigation approach.     
\end{abstract}

\section{Introduction}
Cardiac magnetic resonance imaging (MRI) is an established tool for the diagnosis of various cardiomyopathies~\cite{3033-01,3033-02}. For a comprehensive diagnosis, two factors are essential: First, the anatomy of the heart has to be imaged for the evaluation of different cardiac structures. Second, dynamic imaging, i.e., the resolution into different cardiac phases, is needed for the evaluation of the cardiac function. Recent 3-D protocols were proposed for the sampling of dynamic cardiac 3-D volumes~\cite{3033-03,3033-04,3033-05}. Most protocols sample data continuously during free-breathing and multiple cardiac cycles, often combined with incoherent subsampling~\cite{3033-04,3033-05}. However, the permanent respiration motion during scanning can have a substantial influence on the image quality, yielding severe artifacts. This can be improved by reconstructing data from only one respiration state. Recent approaches for respiration extraction use either additional navigation readouts~\cite{3033-06} (which prolong the scan times), or 1-D self-navigation, where central $k$-space lines (mainly orientated in the superior-inferior (SI) direction) are processed with, e.g., principal component analysis (PCA)~\cite{3033-05,3033-07}. However, only 1-D central $k$-space lines might be insufficient for the extraction of respiration, as it mainly comprises anterior-posterior (AP) as well as SI motion. Consequently, this motion is not properly encoded in the 1-D central $k$-space lines.  In this work, we propose an adapted sampling and navigation framework for extracting 2-D respiration motion from continuous cardiac sequences to reduce the influence of respiration-induced artifacts. We show the superior performance of our 2-D navigation on the image quality compared to no respiration navigation and previous 1-D navigation. 

\section{Method}
Our proposed framework uses 2-D navigation signals from the adapted sampling scheme (Sec.~\ref{3033-sec2-1}, Fig.~\ref{3033-fig1}) for extracting the different respiration states (Sec.~\ref{3033-sec2-2}, Fig.~\ref{3033-fig2}). Readouts from one particular respiration state are jointly reconstructed, reducing respiration-induced artifacts. 

\subsection{Sampling pattern}\label{3033-sec2-1}
\begin{SCfigure}[5][bt]
	\setlength{\figbreite}{0.5\linewidth}
	\caption{Adaption of the sampling pattern (figure adopted from~\cite{3033-08}). Samples on the phase-encoding plane are collected continuously on pseudo-spiral spokes, every $t=1\,s$ the navigation signals are sampled on the $k_{z}=0$, $k_y=-u/2, ..., 0, ..., u/2$ line.}
	\includegraphics[width=0.5\textwidth]{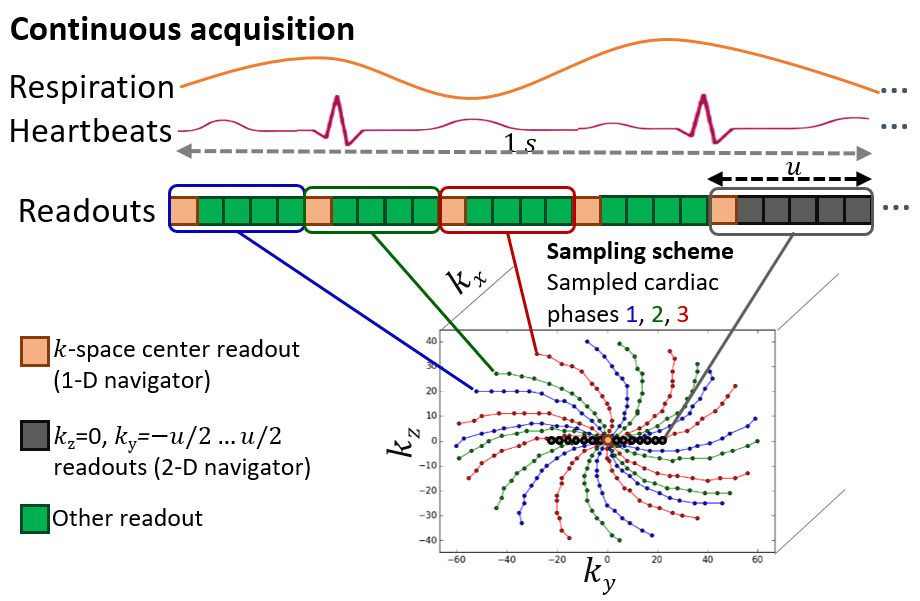}
	\label{3033-fig1}
\end{SCfigure}

\begin{figure}[b]
	\centering
	\includegraphics[width=0.9\textwidth]{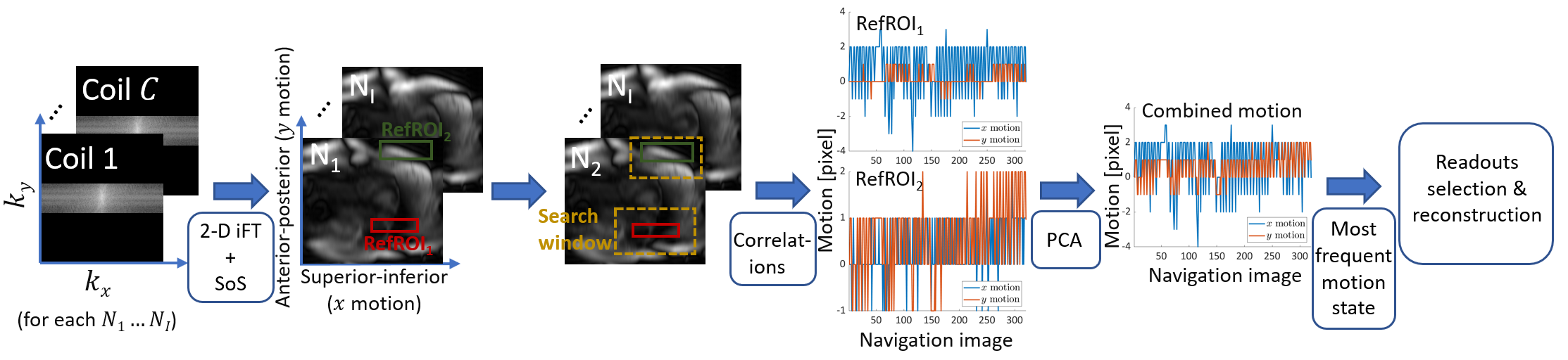}
	\caption{Respiration motion extraction pipeline. Raw navigation signals from single coils are reconstructed with the 2-D inverse Fourier transform (iFT) and sum-of-squares (SoS) coil combination. For motion extraction, correlations between different ROIs on navigation images are computed. PCA extracts the main motion values. The cluster with the highest amount of readouts corresponding to one particular motion state is used for image reconstruction.}
	\label{3033-fig2}
\end{figure}

We extended a previously proposed prototypical method for 3-D free-breathing, continuous, cardiac MRI~\cite{3033-05}. Data is collected during multiple cardiac and respiration phases. The Cartesian phase encoding (PE) plane is incoherently undersampled with samples on pseudo-spiral spokes, each starting in the $k$-space center (Fig.~\ref{3033-fig1}). We adapted this scheme to sample 2-D respiration motion information: Every $t=1\,s$, data is sampled on the $k_{z}=0$, $k_y=-u/2, ..., 0, ..., u/2$ line, where $u$ is number of samples on one pseudo-spiral spoke. This leads to a fully sampled $k$-space center with size $u\times v$, where $v$ is the length of the $k_x$-line, the remaining positions are zero-filled. The 2-D navigation signals are planes within the imaged 3-D volume, mainly orientated in the AP and SI direction. These readouts are simultaneously used for navigation and reconstruction of the 3-D dynamic volumes, avoiding additional scan time for navigation. All acquired readouts after a navigation signal $i$ and prior to the next navigation signal $i+1$ are considered to have the same motion state as the navigation signal $i$. 

\subsection{Navigation motion extraction pipeline}\label{3033-sec2-2}
All 2-D navigation signals from single coils are transformed to the spatial domain with the 2-D inverse Fourier transform (iFT) and are combined with the sum-of-squares (SoS) resulting in final low-resolutional navigation images, $N_{1}, ..., N_{I}$, where $I$ is the total number of navigation images. The 2-D respiration motion in AP and SI direction can be observed there (Fig.~\ref{3033-fig3}). For the selection of one respiration state, following steps are applied: (1) The first navigation image is selected as reference $N_{\text{ref}}$ with $R$ different regions of interest (ROIs) $N_{\text{refROI}_1}, ..., N_{\text{refROI}_R}$ with observable motion. (2) For each ROI the correlation coefficients $CC$ between $N_{\text{refROI}_1}, ..., N_{\text{refROI}_R}$ and $N_{\text{ROI}_1}, ..., N_{\text{ROI}_R}$ in each $N_{2}, ..., N_{I}$ are calculated, while moving the ROI within a selected search window on $N_{2}, ..., N_{I}$. The spatial positions of $N_{\text{ROI}_1}, ..., N_{\text{ROI}_R}$ in $N_{2}, ..., N_{I}$ with the highest $CC$ result in the $x$ and $y$ shifts for the two motion directions compared to the $N_{\text{refROI}_1}, ..., N_{\text{refROI}_R}$, yielding the current motion vector $\vec{m_{ir}}=(\hat{x}_{ir},\hat{y}_{ir})^{\top}$ for each $N_{2}, ..., N_{I}$ and each $N_{\text{ROI}_1}, ..., N_{\text{ROI}_R}$. (3) We assume that each ROI will mainly contain one direction of motion (e.g., a ROI on the chest wall will contain mainly AP motion). PCA is applied on all motion vectors from step (2) to select the dominant 1-D shifts from each ROI, resulting in a combined $\vec{m_{comb}}=(\hat{x}_{ir_{\max}},\hat{y}_{ir_{\max}})^{\top}$ for each $N_{2}, ..., N_{I}$, where $r_{\max}$ is the selected ROI for one particular motion direction ($x$ or $y$). (4) Different motion states and the amount of respective navigation images for each state are computed. All readouts from the state with the most frequent occurence are selected for reconstruction, which is based on a previously proposed prototypical Compressed Sensing framework combined with spatiotemporal Wavelet regularization~\cite{3033-08}.

\section{Experiments and results}
\subsection{Experimental setup}
\begin{figure}[b]
	\centering
	\includegraphics[width=0.7\textwidth]{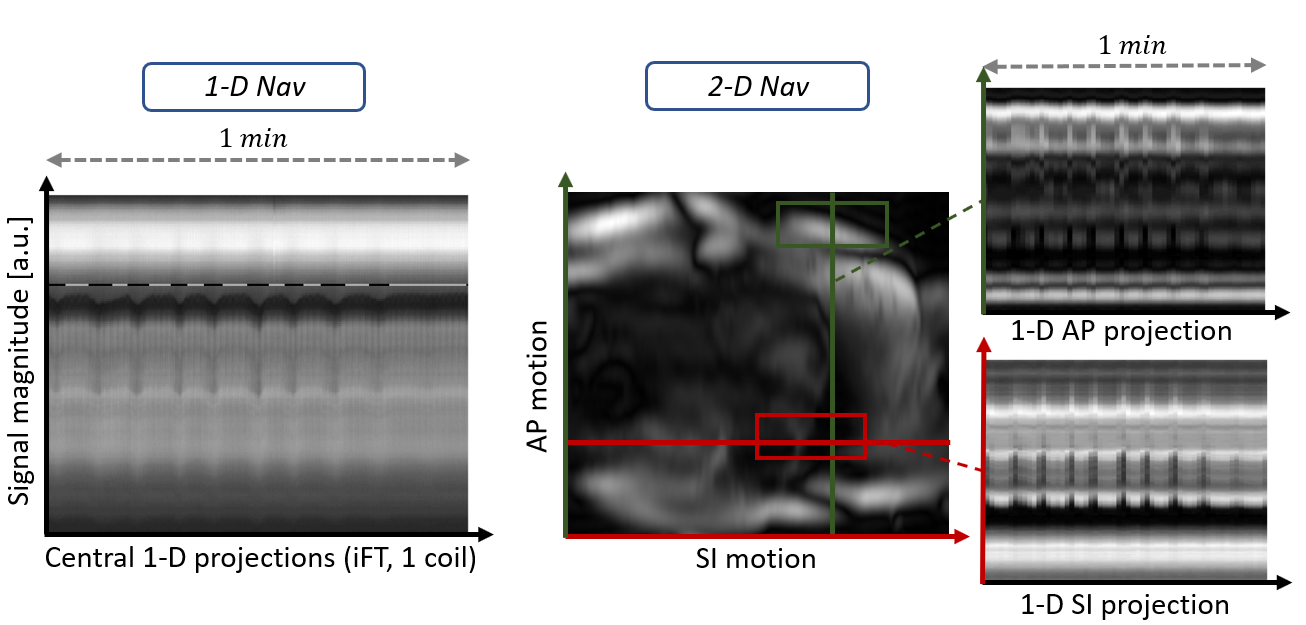}
	\caption{Examples of 1-D lines for extracting 1-D motion from central $k$-space lines (left, readouts corresponding to different states are marked with the black and white line, respectively), and navigation images (right) for extracting 2-D motion from our proposed 2-D navigation, which result in 1-D projections over time, mainly orientated in AP and SI directions.}
	\label{3033-fig3}
\end{figure}

We acquired a 3-D free-breathing, continuous in-vivo scan from one volunteer (female, 55 years) after written consent was obtained. Data was acquired with the proposed prototypical sequence in short-axis orientation and balanced steady-state free precession readouts on a $1.5\,T$ scanner (MAGNETOM Aera, Siemens Healthcare, Erlangen, Germany) with these parameters: Field-of-view: $310\times 310\times86\,mm$, spatial resolution: $1.8^3\,mm$, flip angle: 46\,$^{\circ}$, echo time: $1.7\,ms$, repetition time: $3.3\,ms$, scan time: $5.3\,min$. Cardiac signal from an external ECG device was used to bin the data into cardiac phases~\cite{3033-05}. We selected $R=2$ navigation ROIs, which correspond mainly to AP (green ROI in Fig.~\ref{3033-fig3}) and SI motion (red ROI in Fig.~\ref{3033-fig3}). A search window of $\pm$\,25 pixels in $x$, $y$ directions on $N_{2}, ..., N_{I}$ was used. The $\vec{m_{comb}}$ was selected with $\hat{x}=2$ (from green ROI with a principal component coeffiecent of 0.99) and $\hat{y}=1$ (from red ROI with a principal component coeffiecent of 0.99) with 39.9\,\% of all data (108/339 navigation images). We compared three different reconstructions: (1) Without respiration compensation (\textit{No Nav}), (2) with 1-D navigation (\textit{1-D Nav}), with respiration motion extracted from central 1-D $k$-space lines~\cite{3033-05} (Fig.~\ref{3033-fig3}) and (3) with our proposed 2-D navigation (\textit{2-D Nav}). For the quantitative evaluation, three commonly used image quality metrics in motion compensation~\cite{3033-09} were applied (Tab.~\ref{3033-tab1}): (1) Histogram entropy $H$, (2) total variation $TV$ and (3) wavelet-based estimation of the standard deviation of Gaussian noise distribution $\sigma_{\text{Noise}}$~\cite{3033-10}.

\subsection{Results}
\begin{figure}[b]
	
	\centering
	\resizebox{0.945\textwidth}{!}{
		
			
			\begin{tabular}{ c c c c c c}
				& & \textit{No Nav}& \textit{1-D Nav} & \textit{2-D Nav} & |\textit{No Nav}$-$\textit{2-D Nav}|\\
				& & & & & \includegraphics[width=0.205\textwidth]{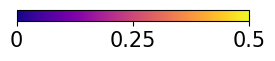}%
				\\
				&
				\rotatebox{90}{\bfseries{2}} 
				&\includegraphics[width=0.205\textwidth]{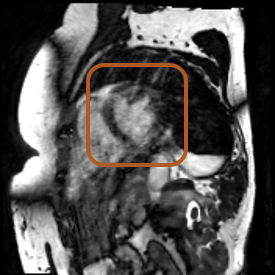}%
				&\includegraphics[width=0.205\textwidth]{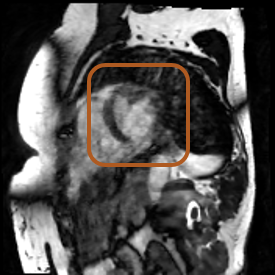}%
				&\includegraphics[width=0.205\textwidth]{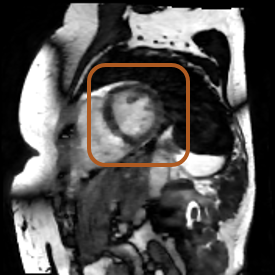}%
				&\includegraphics[width=0.205\textwidth]{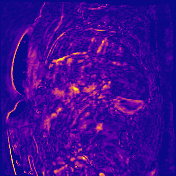}%
				\\
				\rotatebox{90}{\bfseries{Cardiac phases}} &
				\rotatebox{90}{\bfseries{9}} 
				&\includegraphics[width=0.205\textwidth]{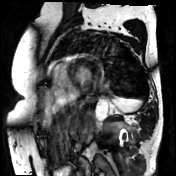}%
				&\includegraphics[width=0.205\textwidth]{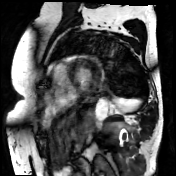}%
				&\includegraphics[width=0.205\textwidth]{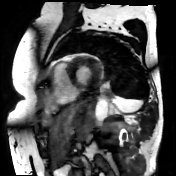}%
				&\includegraphics[width=0.205\textwidth]{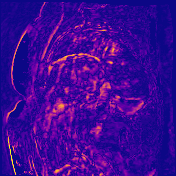}%
				\\
				&
				\rotatebox{90}{\bfseries{14}} 
				&\includegraphics[width=0.205\textwidth]{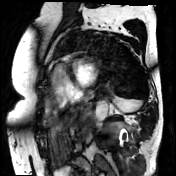}%
				&\includegraphics[width=0.205\textwidth]{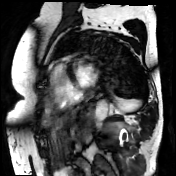}%
				&\includegraphics[width=0.205\textwidth]{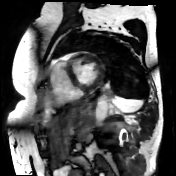}%
				&\includegraphics[width=0.205\textwidth]{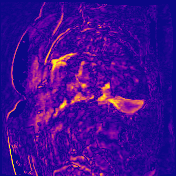}%
				\\
			\end{tabular}%
			
	}
	\caption{Qualitative results. One slice from the reconstructed 3-D volume (48 slices, 20 cardiac phases) is shown for different cardiac phases (each row). Column by column: \textit{No Nav}: Without respiration compensation, \mbox{\textit{1-D Nav}}: With 1-D respiration navigation, \textit{2-D Nav}: With 2-D respiration navigation, |\textit{No Nav}$-$\textit{2-D Nav}|: Difference between reconstructions without navigation and 2-D respiration navigation.}
	
	\label{3033-fig4}
\end{figure}

Figure~\ref{3033-fig4} shows one slice resolved into different cardiac phases from 3-D volumes reconstructed with different approaches. Our 2-D based navigation shows a visually sharper image quality compared to the other reconstructions, especially in the cardiac region (marked with orange box for cardiac phase 2). For all quantitative metrics shown in  Tab.~\ref{3033-tab1}, the 2-D based navigation approach yields superior results. The Student's $t$-test results confirm that the differences in the image quality between the approaches are statistically significant ($p$-value\,$<0.05$).

\begin{table}[t]
	\caption{Quantitative image quality metrics for different reconstructions, each given as $\mu\pm\sigma$ for the 3-D dataset (48 slices, 20 cardiac phases). \textit{No Nav:} Without respiration compensation, \textit{1-D Nav:} With 1-D navigation~\cite{3033-05}, \textit{2-D Nav:} With the proposed \mbox{2-D} navigation, $H$: Histogram entropy (lower is better), $TV$: Total variation (lower is better), $\sigma_{\text{Noise}}$: Standard deviation of Gaussian noise distribution (lower is better).}
	\label{3033-tab1}
	\begin{tabular*}{\textwidth}{l@{\extracolsep\fill}c c c c}
		\hline
		Method & \multicolumn{3}{c}{Metric ($\mu\pm\sigma$), [95\,\% confidence intervals]}\\
		&$H$ & $TV$ & $\sigma_{\text{Noise}}$\\ 
		\hline
		\textit{No Nav} & 3.80\,$\pm$\,0.11\,[3.79, 3.80]& 90868\,$\pm$\,790\,[90817, 90918] & 6.17\,$\pm$\,1.00\,[6.11, 6.24]\\
		\textit{1-D Nav} & 3.77\,$\pm$\,0.12\,[3.77, 3.78] & 90575\,$\pm$\,856\,[90521, 90629] &5.41\,$\pm$\,0.95\,[5.35, 5.47]\\
		\textit{2-D Nav} & \bfseries{3.68}\,$\pm$\,0.11\,[3.67, 3.68] & \bfseries{86197}\,$\pm$\,1831\,[86081, 86313] &  \bfseries{4.36}\,$\pm$\,0.69\,[4.31, 4.39]\\ 
		\hline
	\end{tabular*}
\end{table}

\section{Discussion and conclusion}
No respiration compensation yields artifact-corrupted images, which are of limited diagnostic utility: The myocardial structure (marked with orange box for cardiac phase 2) is heavily disturbed by the permanent chest wall motion (especially in cardiac phases 2, 14), that manifests as folding artifacts. The reconstruction based on the 1-D navigation also results in artifact-corrupted images. Even if 1-D motion is visible in the central 1-D $k$-space lines (Fig.~\ref{3033-fig3}), it can only represent one motion direction, which is insufficient for respiration. For our subject, the image quality could only be slightly improved with the 1-D navigation. Our proposed 2-D navigation yields the visually sharpest image quality. Even if the subsampling is increased with the 2-D navigation by removing parts of data for the reconstruction (up to a factor 11.5, compared to max. 4.1 without respiration compensation), the image quality can be greatly improved with data from only one stable respiration state. This is confirmed by our quantitative analysis, where the 2-D navigation has significantly superior results in all metrics. 

To summarize, we proposed a sampling pattern adaption and a 2-D respiration navigation pipeline that improves image quality compared to previously proposed 1-D navigation. Using our adapted continuous sampling, the same data can be used for 2-D navigation and \mbox{3-D} reconstruction, without introducing the need for sampling navigation data only. Our approach can be potentially combined with a data-driven cardiac navigation, e.g.,~\cite{3033-11} without further adaptions on the sampling, yielding an ECG-free workflow. Future work will deal with the automatic detection of ROIs for the motion computation, as well as the application to other protocols, such as multi-contrast sampling~\cite{3033-05}.

\bibliographystyle{bvm}

\bibliography{3033}

\marginpar{\color{white}E\articlenumber}

\end{document}